\documentclass{Interspeech}
\interspeechcameraready

\title{Towards disentangling the contributions of articulation and acoustics in multimodal phoneme recognition}

\author[affiliation={1,2}]{Sean}{Foley}
\author[affiliation={1}]{Hong}{Nguyen}
\author[affiliation={1}]{Jihwan}{Lee}
\author[affiliation={1}]{Sudarsana Reddy}{Kadiri}
\author[affiliation={2}]{Dani}{Byrd}
\author[affiliation={2}]{Louis}{Goldstein}
\author[affiliation={1}]{Shrikanth}{Narayanan}

\affiliation{Signal Analysis and Interpretation Laboratory}{University of Southern California}{USA}
\affiliation{Department of Linguistics}{University of Southern California}{USA}
\email{seanfole@usc.edu}

\usepackage{comment}
\usepackage{graphicx}      
\usepackage{subcaption} 
\usepackage{tipa}

\begin{document}

\maketitle

\begin{abstract}
Although many previous studies have carried out multimodal learning with real-time MRI data
that captures the audio-visual kinematics of the vocal tract during speech, these studies have been limited by their reliance on multi-speaker corpora. This prevents such models from learning a detailed relationship between acoustics and articulation due to considerable cross-speaker variability. In this study, we develop unimodal audio and video models as well as multimodal models for phoneme recognition using a long-form single-speaker MRI corpus, with the goal of disentangling and interpreting the contributions of each modality. Audio and multimodal models show similar performance on different phonetic manner classes but diverge on places of articulation. Interpretation of the models' latent space shows similar encoding of the phonetic space across audio and multimodal models, while the models' attention weights highlight differences in acoustic and articulatory timing for certain phonemes.
    
\end{abstract}

\section{Introduction}
Speech is inherently multimodal, with the integration of sensory and motor information being essential to both speech production and perception \cite{hickok2011sensorimotor, venezia2016perception, goldstein2019role}. In speech production, individual variability in both articulation and its consequent acoustics is robust \cite{whalen2018variability, harper2021individual}. A crucial aspect of this variability concerns the realization of targets, even within the long-standing debates as to whether the targets of speech production are acoustic or articulatory or a combination of both \cite{browman1992articulatory,hickok2014architecture, faytak2018articulatory}. In continuous speech, achieving a target can be framed as the successful encoding of phonological information, particularly the identity of the phonemes being spoken (in contrast to others). The recognition of phonemes is then a problem of multimodal learning, with the possibility that, for a given phoneme, one modality may be more informative than the other or that both are flexibly integrated at short and/or longer timescales in perception.
\par
In multimodal machine learning, audio-visual models often use video that captures the movements of the jaw, face and lips \cite{shi2022learning, wu2024robust}, a view typical of a listener in everyday conversation. The focus of the current study is multimodal learning using speech data captured using real-time MRI (rtMRI), which typically captures simultaneous video of the soft-tissue vocal tract (generally in mid-sagittal orientation) covering the entire airway from lips to the larynx and speech audio  \cite{lim2024speech,lim2021multispeaker}. Such video allows for capturing the kinematics of the tongue, jaw, lips, velum, and larynx, i.e. the major articulators employed in speech production. This permits models to be better-aligned with the causal mechanisms underlying speech production. 
\par
A number of previous studies have used such speech data in a variety of speech classification tasks \cite{saha2018towards, van2019cnn, kose2021multimodal, yue2024towards, shi2024direct}. \cite{saha2018towards} used solely rtMRI video of the vocal tract to classify different vowel-consonant-vowel (VCV) sequences using a combination of RNNs and CNNs. \cite{kose2021multimodal} used both video and audio to perform phone classification and recognition on the USC-TIMIT corpus \cite{narayanan2013usc}, finding that multimodal information resulted in better performance than either unimodal case. Both \cite{van2019cnn} and \cite{yue2024towards} used the USC Vocal Tract Morphology corpus \cite{sorensen2017database} for phone classification and VCV classification respectively, with the former using a CNN-based approach and the latter a Transformer-based model. With the USC 75-Speaker Dataset \cite{lim2021multispeaker}, \cite{shi2024direct} related self-supervised model performance in phoneme recognition to cross-speaker differences in articulation, as illuminated in MRI video. 
\par
While previous studies have conducted multimodal speech classification using rtMRI audio and video, few have specifically attempted to disentangle the contribution of each modality in the given task. Typically, differences in some metric across unimodal audio/video only and multimodal models are used to assess the benefit of multimodal information 
\cite{kose2021multimodal,yue2024towards}. Crucially, this method primarily serves to distinguish model performance but does not offer a robust measure of the information gain provided by each modality, particularly in the identification of individual phonemes. Additionally, all previous studies have relied on multimodal data from multiple speakers. This limits the model's ability to learn a concise and veridical relationship between acoustics and articulation, given the considerable cross-speaker variability in articulation and vocal tract morphology \cite{whalen2018variability, serrurier2019characterization}.  
\par

\par
The current study aims to disentangle the role of acoustics and articulation in the task of phoneme recognition from continuous speech, with a particular focus on the benefits gained by adding articulatory information in comparison to using audio features alone. The study makes use of a long-form single speaker rtMRI corpus, allowing the models to learn a robust relationship between acoustic and articulatory information. 
The results highlight similarities and differences in the performance of audio-only and multimodal models in identifying phonemes of certain phonetic classes. Additionally, the current study develops tactics for interpreting both the latent space and the attention weights of such models. 
As such, this study provides one of the more thorough interpretations of multimodality in phoneme recognition.

\section{Method}

\subsection{Model}
The primary module for the model architecture is a Conformer, following the design in \cite{gulati2020Conformer}, which consists of two feed-forward modules, a self-attention layer, and a convolution module (see \cite{gulati2020Conformer} for further details), accessed via Torchaudio. 
The output of the Conformer is decoded by a single LSTM layer, with a final linear layer used for prediction. In the unimodal cases, the input to the Conformer consists of either acoustic features $\textbf{A} \in \mathbb{R}^{t \times D}$ or video features $\textbf{V} \in \mathbb{R}^{t \times D}$, where $t$ is the temporal dimension and $D$ is the feature dimension. For the multimodal model, the audio and video features are concatenated along the temporal dimension, such that $\textbf{M} \in \mathbb{R}^{t \times 2D}$. 
For the task of phoneme recognition, the model is trained using the connectionist temporal classification (CTC) loss \cite{ctc-loss}. 
As a baseline, we performed zero-shot phoneme recognition on the audio using Wav2Vec2Phoneme, accessed via Hugging Face \cite{xu2021simple}.

\begin{figure}
    \centering
    \includegraphics[width=\linewidth]{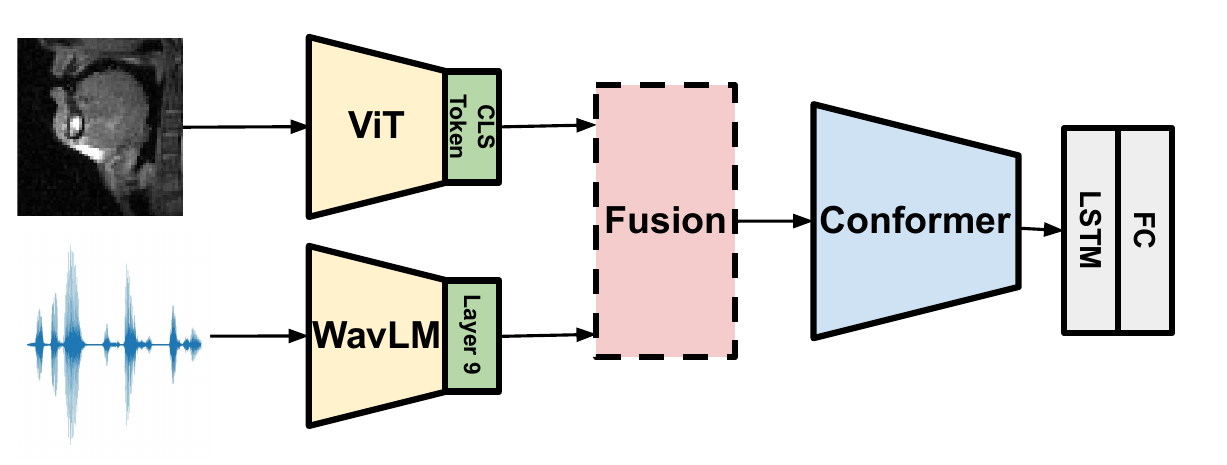}
    \caption{Model architecture for the current study.}
    \label{fig:modelarch}
\end{figure}

\subsection{Analysis}

Overall phoneme recognition is calculated using Phoneme Error Rate (PER). Additionally, PER for individual phonemes is calculated using the edit distance for sequences containing the target phoneme and dividing by the number of occurrences of that phoneme in the ground truth sequence. These values were averaged across manner and place of articulation classes. To get a sense of how acoustic, video, or multimodal features may capture the phonetic space differently, the latent representations from the Conformer convolution module were extracted for individual phonemes by slicing them in the temporal dimension using the ground truth time-stamps and then averaging across the temporal dimension.
\par
Following \cite{yue2024towards}, the attention weights from the self-attention layer of the last Conformer layer were  extracted. While \cite{yue2024towards} considered the effects of adding audio features to the attention weights in comparison to using only video features, we investigate the opposite effect, specifically the effect of adding video features compared to using only audio features. This tactic stems from the possibility that differences in articulatory and acoustic timing may be evident for certain phonemes. The attention weights were averaged across the attention heads, and then for each key the average query value was calculated, which provides values for the degree to which each time step is attended to by the entire sequence. Phoneme time stamps were used to get phoneme-level attention values, with 25 ms windows on either side to capture potential coarticulatory effects. All attention weights were $z$-scored after averaging across the attention heads. 

\section{Experiments}

\subsection{Dataset}

We use a new single speaker rtMRI corpus with simultaneously recorded audio and video. The corpus contains speech data from one male native speaker of American English producing the 460 sentences used in the USC TIMIT corpus, passages used in the USC 75-Speaker Dataset \cite{lim2021multispeaker} and spontaneous speech prompted on topics such as food, hobbies, travel, etc. The vocal tract of the speaker was imaged in midsagittal orientation using a 0.55T MRI scanner with a custom upper airway receiver coil \cite{munoz2023evaluation}, resulting in a frame rate of 99 frames/sec \cite{kumar24b_interspeech}. Audio was collected at 16 kHz. In total, approximately one hour of audio data was collected, comparable to the MNGU0 EMA corpus \cite{richmond2011announcing}. First-pass phoneme alignments were extracted using the Montreal Forced Aligner (MFA) and manually corrected by a phonetician in Praat. 

\subsection{Pre-processing}

\textbf{Audio:} From the raw audio, speech segments were extracted using the pre-trained Voice Activity Detection (VAD) model pyannote, accessed via Hugging Face. The resulting VAD speech segments were concatenated or split into chunks of up to 5 sec in length. Word boundaries were used to split speech segments longer than 5 sec. Both the audio and video were chunked using these time intervals, resulting in 767 chunks, for a total duration of 38 min and an average duration of 2.95 sec per chunk. These chunks were split into train/test sets with a ratio of 0.7/0.3. The audio chunks were first denoised using the denoiser model in \cite{defossez2020real}, and then features were extracted from the $9^{th}$ layer of the pre-trained base WavLM model \cite{chen2022wavlm}, based on previous work demonstrating that this layer encodes rich phonetic information \cite{cho2022evidence}.

\noindent
\textbf{Video:} The original MRI video was cropped to focus on the vocal tract region, with the upper hard palate, larynx, and pharyngeal wall serving as boundaries (see Figure~\ref{fig:modelarch}). All video frames were rescaled to a spatial resolution of 224 $\times$ 224. For each frame, we took the classification (CLS) token from the last hidden layer of the base ViT model \cite{alexey2020image},
which was fine-tuned on MRI video from 10 native American English speakers in the USC 75-Speaker Dataset \cite{lim2021multispeaker} using a self-supervised approach \cite{simclr, azizi2022robust}. 
Linear interpolation was used to match the sampling rate of the audio and video features, with the audio features being upsampled. 

\subsection{Implementation details}

The Conformer model was initialized with the same general configuration described in \cite{gulati2020Conformer}. The input dimension to the Conformer was set to 768 for the unimodal models and 2*768 for the multimodal model, with a feed-forward dimension of 256. The number of attention heads was set at 4, a kernel size of 31, and dropout set at 0.3. The number of Conformer layers was set to 3. For the LSTM layer, the latent size was 128, which served as the input size for the final linear layer. For all models, the Adam optimizer was used, with a batch size of 8 and a learning rate of 1e-3, which was decayed by a factor of 0.9 every 20 epochs. Learning rate, batch size, and weight decay were set using a grid search over the range of [1e-3, 5e-4, 1e-4], [8, 16, 32], and [1e-3, 5e-4, 1e-4] respectively. All models were trained on an NVIDIA A40 GPU.

\section{Results}

\subsection{Phoneme error rate (PER)}

The overall PER results on the held-out test set are presented in Table~\ref{tab:two_column_table}. The results reported here are from the best performing model on the test set, based on the CTC loss. The baseline Wav2Vec2Phoneme model achieved a zero-shot PER of 0.36. While no previous study has performed phoneme recognition on the corpus used here, we include as a further comparison the results from \cite{kose2021multimodal}, which is one of the more recent studies that carried out multimodal phoneme recognition on a similar rtMRI video and audio corpus, the USC TIMIT corpus \cite{narayanan2013usc}. As can be seen, our audio and multimodal models greatly outperform the baseline comparisons, achieving PERs of 0.21 and 0.26 respectively. The video model only achieves a PER of 0.49. 

\begin{table}[h!]
\centering
\caption{PER results for models in current study (bottom) compared to baseline models (top), where K- refers to the results from \cite{kose2021multimodal}. Models from the current study are in bold.}

\begin{tabular}{cc}
\hline
\textbf{Model} & \textbf{PER} \\ \hline
Wav2Vec2Phoneme            & 0.36           \\ 
K-Audio            & 0.40           \\ 
K-Video            & 0.46        \\ 
K-Multimodal            &   0.38         \\
\midrule
\textbf{Audio}            & \textbf{0.21}           \\ 
\textbf{Video}            & \textbf{0.49}        \\ 
\textbf{Multimodal}            &   \textbf{0.26}          \\ \hline
\end{tabular}
\label{tab:two_column_table}
\end{table}

\begin{figure}
    \centering
    \includegraphics[width=\linewidth]{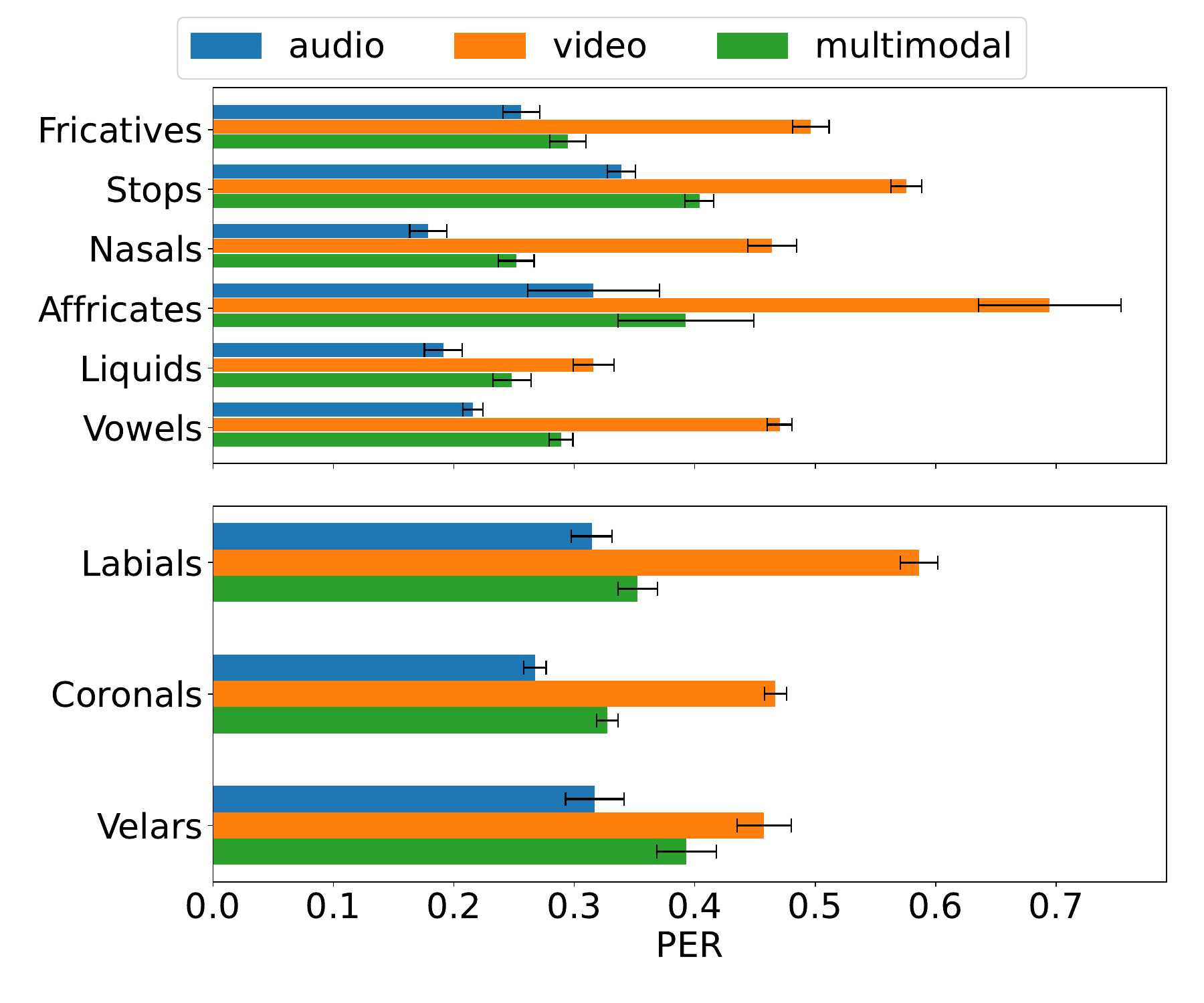}
    \vspace{-0.7cm}
    \caption{PER results by phonetic class for models in the current study, grouped by manner (above) and place of articulation (below).}
    \label{fig:phoneme-per}
\end{figure}

The PER results per phoneme class are reported in Figure~\ref{fig:phoneme-per}, with results shown for unimodal models and the multimodal model. Bootstrap sampling was used to compute error rates, using 1,000 iterations. The top section is split into manner classes and the bottom section is grouped by place of articulation. All models perform best on nasals, liquids, and vowels and worst on fricatives, stops, and affricates. Affricate recognition is especially poor for the video model, and the audio model performance on this class is not significantly better than the multimodal model. Notably, the relative performances on the six manner classes is comparable across the three models. 

For place of articulation, the cross-model performances vary more. While audio performs best on coronals, video does slightly better on velars than coronals. The video model performance on labials is considerably worse, while the audio model performs comparably on labials and velars. Interestingly, the multimodal model PER is worst on velars and best on coronals. These discrepancies highlight that for some places of articulation multimodality can be additive and for others subtractive.

\subsection{Interpretation of latent space}

\begin{figure}[!ht]
    \centering
\includegraphics[width=.49\linewidth]{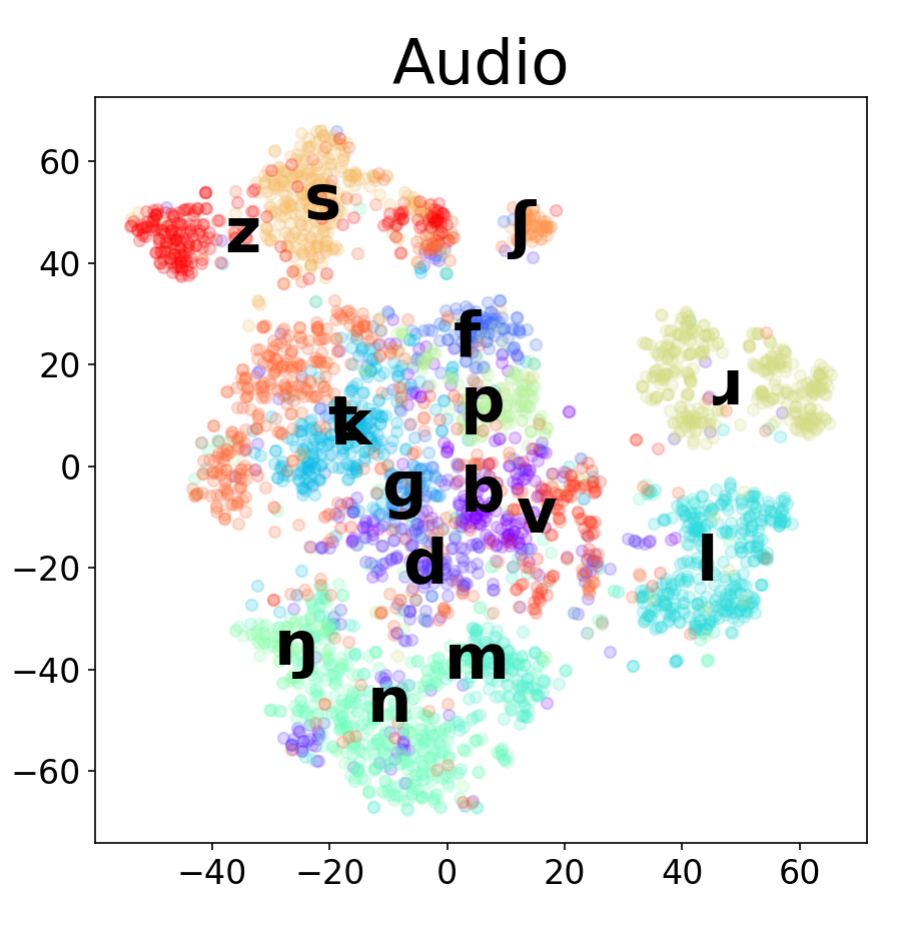}\hfill
\includegraphics[width=.49\linewidth]{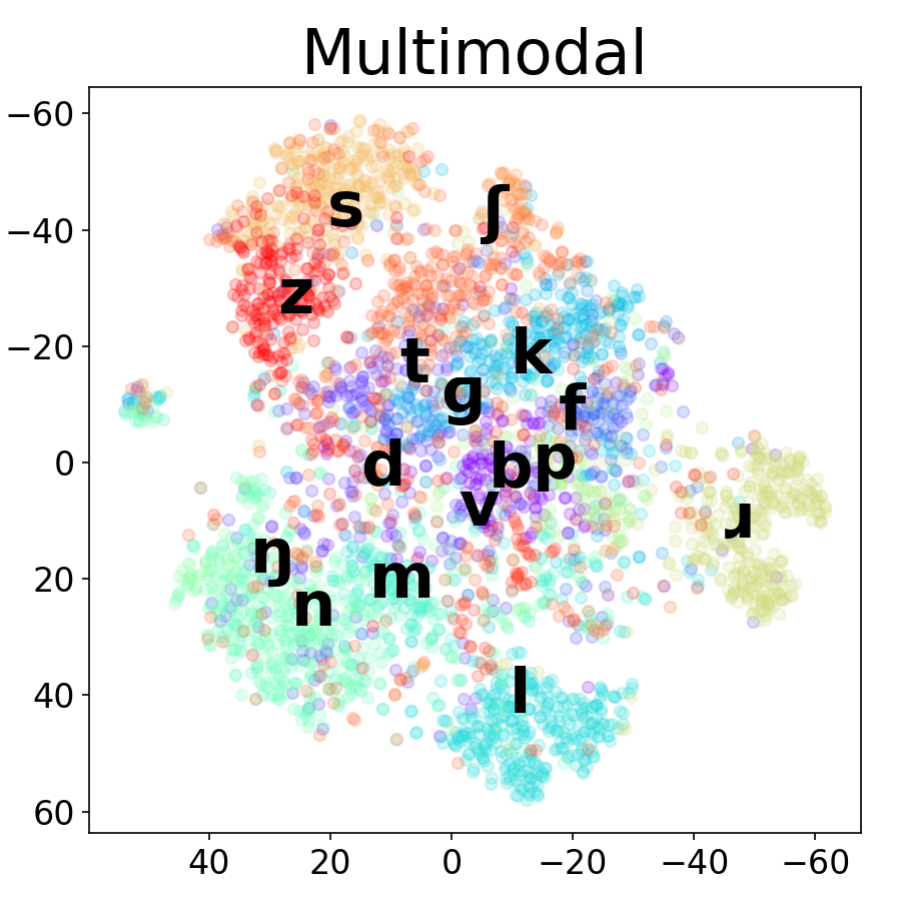}
\caption{t-SNE plot of the latent space for consonants for unimodal audio (left) and multimodal (right) models.}
\label{fig:tsne}
\end{figure}

The phoneme-level representations were concatenated across all items, grouped by consonants and vowels, and submitted to a t-SNE \cite{van2008visualizing} to project the representations to a 2D space, with perplexity set to 30. The results for consonants are shown in Figure~\ref{fig:tsne}. It can be seen clearly that both models learn distinct representations for the liquids, nasals and sibilants /s z \textesh/. Where the representations are less distinct is in the oral labial and coronal stops and labio-dental fricatives, which show a great deal of overlap. Also, in the audio feature space, the phonemes /t/ and /k/ have centroids in nearly the same position, despite these phonemes having different places of articulation, with similar overlap seen for the voiced  stops /b d g/. This suggests that the audio model has some difficulty in picking up the frication generated by the labio-dental fricatives and that in like voicing classes the formant transitions that can serve as cues to distinguish places of articulation in stops are not particularly salient. In addition to the inherent subtlety of these distinctions in the audio domain, contextual effects in running speech or the denoising of the audio could also contribute to the indiscreteness of frication noise or formant information. 

The multimodal model shows better distinction between /t/ and /k/; though in general there do not appear to be uniform changes in the latent space for consonants when adding video features. The vowel feature spaces (not shown) show similar clusters across the two models, with most of the peripheral vowels showing well-formed clusters; though the vowels /\textscripta/ and /\textopeno/ have a great deal of overlap, suggesting that the speaker exhibits the \textit{cot-caught} merger. 

\subsection{Interpretation of attention weights}

\begin{figure}
    \centering
    \includegraphics[width=\linewidth]{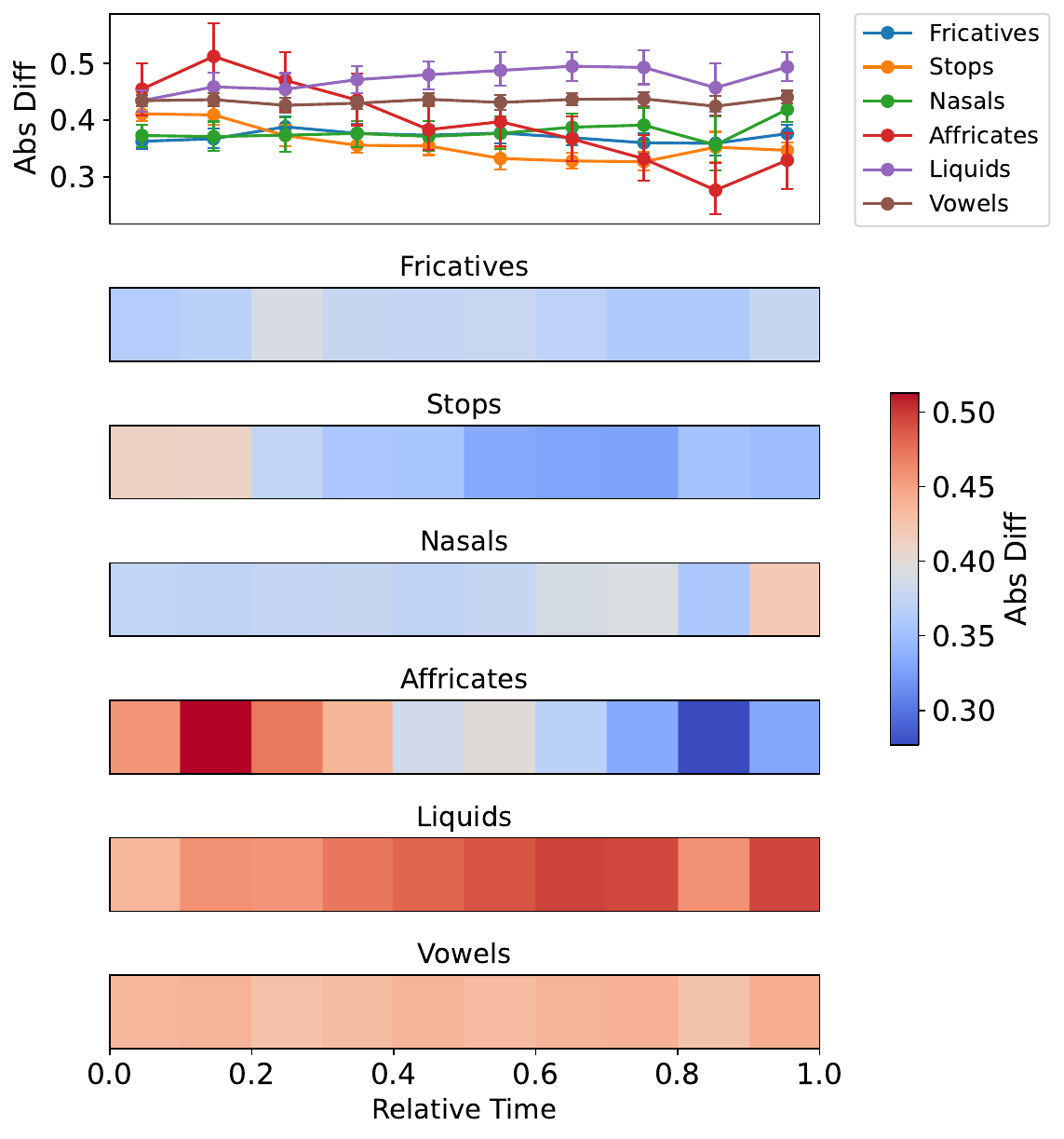}
    \caption{Absolute difference in attention weights at each time step between audio and multimodal models per phoneme manner class.}
    \label{fig:abs-diff}
\end{figure}

For each time stamp in the phoneme-level attention weights, the absolute difference between that weight in the audio and in the multimodal model was calculated and averaged across the manner classes. Ten time bins were used to relativize time across the phonemes. This can be seen in Figure~\ref{fig:abs-diff}, where the top panel shows the average absolute difference between the model weights at each timepoint for each manner class, with bootstrap sampling used to calculate error rates, and the lower panels are heatmaps showing these differences for each manner class. A general trend here is that the differences are greater towards the temporal edges, likely due to coarticulation being more evident in the video. Liquids and vowels overall have the greatest weight differences among the classes and this is true throughout their intervals. The greatest overall difference during the onset of affricates may pertain to the sequential stop-to-fricative transition of these segments. 
\par
\begin{figure}
    \centering
    \includegraphics[width=\linewidth]{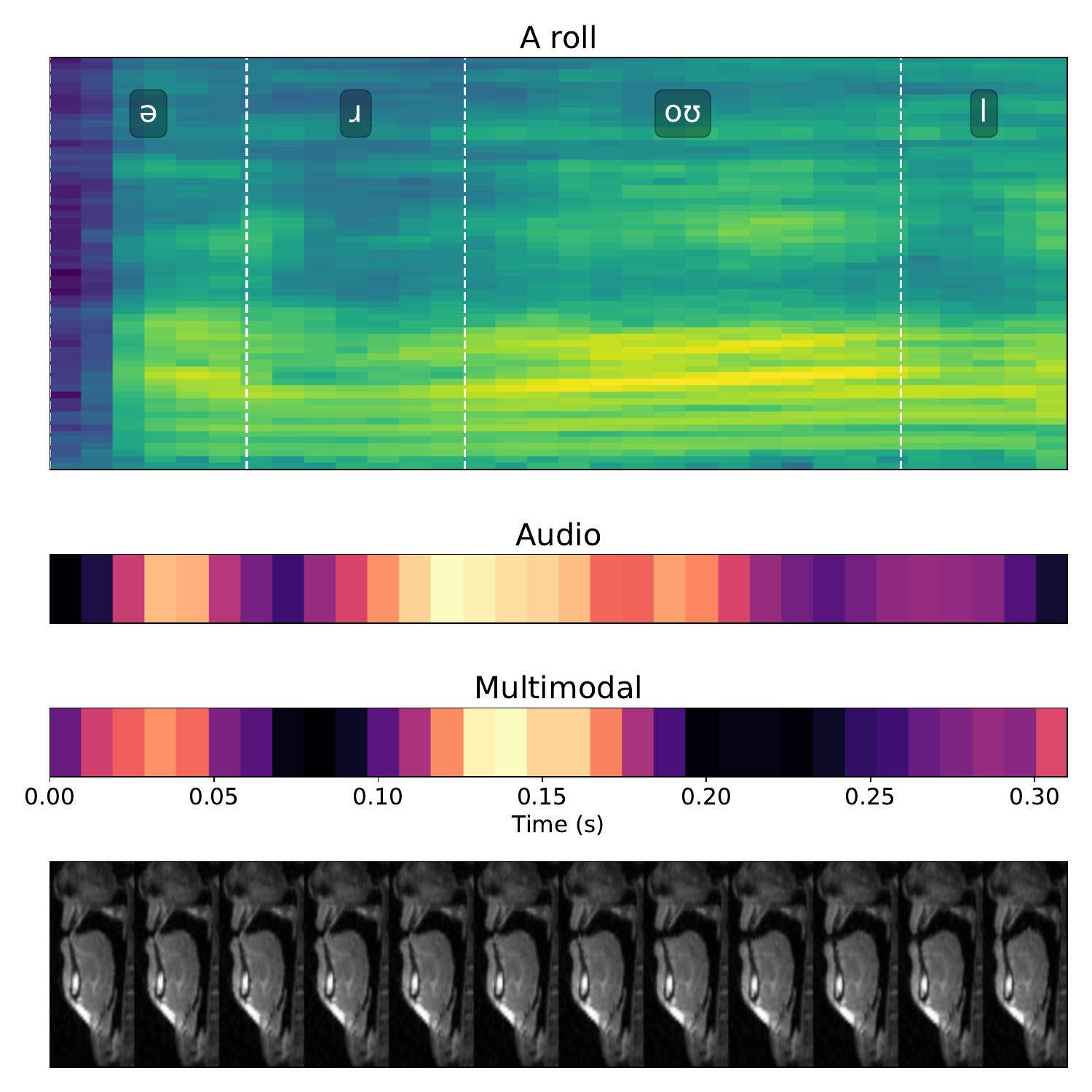}
    \caption{Log-mel spectrogram (top) and audio and multimodal attention weights (middle) for the phrase ``a roll", with MRI frames for the phonemes /\textschwa/ and /\textturnr/ (bottom). Lighter attentions have higher values.}
    \label{fig:att-vis}
\end{figure}

An example highlighting the timing differences across the modalities is shown in Figure~\ref{fig:att-vis} for the phrase ``a roll". Notice that there is quite a discrepancy between the two models during the /\textschwa/, /\textturnr/, and /o\textupsilon/ intervals. The bottom panel shows the MRI video frames during /\textschwa/ and /\textturnr/. The multimodal model attends more to the onset of the /\textschwa/ interval and more towards the end of the /\textturnr/ interval, with clear unattended regions after these times. The audio model attends more to the center of /\textschwa/ and overall much more to the /\textturnr o\textupsilon/ sequence. The MRI frames show clearly that the tongue tip constriction for /\textturnr/ starts to form \textit{before} the acoustic onset of /\textschwa/, while the /o\textupsilon/ constriction is formed during the /\textturnr/ offset. While the audio model attention aligns more with the formant structure, the multimodal model attention is highly localized to intervals containing crucial constriction information. 



\section{Discussion}
In this study, we trained unimodal audio and video models and a combined multimodal model on phoneme recognition using a long-form single-speaker rtMRI corpus. The PER results showed that our audio model significantly outperforms the baseline model and those from similar previous work, which attests to the robustness of the WavLM representations in learning phonetic information. As expected, the unimodal video model does not perform as well as the audio model, given that acoustic features generally outperform the more sparse visual speech information in speech recognition tasks. The multimodal model performs only slightly worse than the audio model, suggesting overall that adding the video features does not improve performance. 
\par
At the level of specific phoneme classes, the unimodal audio and video models performed similarly, in that they performed best on nasals and liquids and did not perform as well on affricates and stops. This is unsurprising as nasals and liquids both have distinctive acoustic and articulatory characteristics, such as anti-resonance structure, multiple constrictions, and for rhotics a low third formant \cite{alwan1997toward,proctor2019articulatory}. Capturing the sequential stop-to-fricative transition that distinguishes stops and affricates is likely challenging for both audio and video modalities.
\par
Interpretation of the models' latent spaces aligns with the phoneme-level PER performance in that more well-formed clusters were found for sibilants, nasals, and liquids, while stops and labio-dental fricatives show more local overlap in their clusters. Interestingly, despite all models performing best on liquids, vowels, and nasals, the models' attention weights differed most between the audio and multimodal models for liquids and vowels. This suggests that the addition of video features can actually force the model to attend to different information. Coarticulation in the articulatory domain is robust and may not always have a clear acoustic correlate. As shown in Figure~\ref{fig:att-vis}, it appears likely that phoneme-relevant information is available \textit{earlier} when articulatory information is present and that the multimodal model relies more on this information than the acoustic-only model. 
\par
Despite the progress made here in interpreting multimodality in phoneme recognition, there are a number of limitations to the current study. First, while the corpus used here has more single-speaker MRI data than is available in public datasets, the total speech time after applying VAD was only 38 minutes. Such limited data undoubtedly still limits the ability of the model to learn strong connections between acoustics and articulation. Second, the current study only implemented this approach with an off-the-shelf Conformer architecture, and replicating this work with other architectures for the task of phoneme recognition could offer further insights into the sort of information gained by the incorporation of audio and video (articulation) modalities for this task.


\section{Acknowledgments}
This work was supported by NIH grant T32 DC009975.
\bibliographystyle{IEEEtran}
\bibliography{My_Library}

\end{document}